\documentclass{article}

\usepackage{microtype}
\usepackage{graphicx}
\usepackage{subfigure}
\usepackage{algorithmic}
\usepackage{booktabs} 

\usepackage{hyperref}

\usepackage[accepted]{icml2024}

\usepackage{amsmath}
\usepackage{amssymb}
\usepackage{mathtools}
\usepackage{amsthm}

\usepackage[capitalize,noabbrev]{cleveref}

\theoremstyle{plain}

\theoremstyle{definition}

\theoremstyle{remark}

\usepackage[textsize=tiny]{todonotes}

\usepackage{pifont}

\newcommand{\SLOT}[1]{{CoBalT}}
\DeclareMathOperator*{\argmax}{arg\,max} 
\DeclareMathOperator*{\softmax}{soft\,max} 
\DeclareMathSymbol{*}{\mathbin}{symbols}{"03} 
\DeclareMathSymbol{\ast}{\mathbin}{symbols}{"03}

\usepackage{bbm}

\icmltitlerunning{Unsupervised Concept Discovery Mitigates Spurious Correlations}

\begin{document}

\twocolumn[
\icmltitle{Unsupervised 
Concept Discovery 
Mitigates Spurious Correlations}



\icmlsetsymbol{equal}{*}

\begin{icmlauthorlist}
\icmlauthor{Md Rifat Arefin}{yyy,sch1}
\icmlauthor{Yan Zhang}{comp}
\icmlauthor{Aristide Baratin}{comp}
\icmlauthor{Francesco Locatello}{sch2}
\icmlauthor{Irina Rish}{yyy,sch1}
\icmlauthor{Dianbo Liu}{sch3}
\icmlauthor{Kenji Kawaguchi}{sch3}
\end{icmlauthorlist}


\icmlaffiliation{yyy}{Mila-Quebec AI Institute, Canada}
\icmlaffiliation{comp}{Samsung - SAIT AI Lab, Montreal, Canada}
\icmlaffiliation{sch1}{University of Montreal, Canada}
\icmlaffiliation{sch2}{Institute of Science and Technology Austria}
\icmlaffiliation{sch3}{National University of Singapore}

\icmlcorrespondingauthor{Md Rifat Arefin}{rifat.arefin@mila.quebec}

\icmlkeywords{Machine Learning, ICML}

\vskip 0.3in
]



\printAffiliationsAndNotice{}  

\begin{abstract}
Models prone to spurious correlations in training data often produce brittle predictions and introduce unintended biases. Addressing this challenge typically 
involves methods relying on prior knowledge and group annotation to remove spurious correlations, 
which may not be readily available in many applications.
In this paper, we establish a novel connection between unsupervised object-centric learning and mitigation of spurious correlations. Instead of directly inferring subgroups with varying correlations with labels, 
our approach focuses on discovering \emph{concepts}: discrete ideas that are shared across input samples. Leveraging existing object-centric representation learning, we introduce \SLOT{}: a concept balancing technique that effectively mitigates spurious correlations without requiring human labeling of subgroups.
 Evaluation  across the benchmark datasets for sub-population shifts demonstrate  superior or competitive performance compared state-of-the-art baselines,  without the need
for group annotation. Code is available at \href{https://github.com/rarefin/CoBalT}{https://github.com/rarefin/CoBalT}
\end{abstract}    
\section{Introduction}
\label{sec:intro}
A critical concern with deep learning models 
arises from their well-known tendency
to base their predictions on correlations present in the training data rather than robustly informative features \cite{arjovsky2019invariant, Sagawa2020Distributionally}.   
For instance, in image classification, translating an image by a few pixels \cite{azulay2019why} or modifying the background \cite{Beery_2018_ECCV} 
can drastically change the predictions of the model. 
Often viewed as resulting from the so-called  `simplicity bias' of deep neural networks in the literature~\cite{shah2020pitfalls}, this phenomenon pervades the landscape of deep learning models \cite{Geirhos2020_shortcut}. While models relying on spurious correlations may perform well on average across i.i.d. test data, they often 
struggle on specific subgroups where these correlations do not hold. Common approaches involve partitioning the training data based  on prior knowledge of spurious information and adjusting  the training process to  ensure consistency across these groups \cite{Sagawa2020Distributionally, kirichenko2022last, arjovsky2019invariant}.
However, most real-world datasets lack explicit annotations highlighting spurious information. 
Manual annotation 
is expensive 
and can be ill-defined, as the appropriate groupings may not be immediately apparent. 

On the other hand, self-supervised learning \citep{chen2020simple, caron2020unsupervised, Caron2021EmergingPI, Grill2020BootstrapYO, he2020momentum} has produced powerful representation learners.  Several
methods  \citep{Picie-2021, wen2022self} aim to learn high-level concepts by semantic grouping of areas within an input image into object-centric instances. \citet{wen2022self}, for instance,   
 leverage slot attention~\cite{locatello2020object} to decompose complex scenes into constituent objects via contrastive learning alone. 
While their original aim was downstream task representation learning, we posit that such decomposition can help mitigating spurious correlations.
By treating semantic groupings as concept sources discovered by the model, they can serve as data-driven proxies of subgroup labels.
This differs from existing work in spurious correlation, which typically  directly infers subgroups  (see related work in \cref{sec:related_works}).
Our approach  models concepts that do not necessarily correspond directly to subgroups; typically, we use a significantly larger number of concepts than annotated subgroups in the dataset.

This paper demonstrates the use of object-centric representation learning approaches to design classifiers robust to spurious correlations without the need for human-labeled subgroup annotations.  
We introduce \SLOT{}, a method combining concept discovery with concept balancing for robust classification. \SLOT{} follows a two-stage procedure common in the literature: first, inferring information about the training data, and then  leveraging this information for robust training.  

\begin{figure}[htb]
\centering
\begin{subfigure}
  \centering
  \includegraphics[scale=0.37]{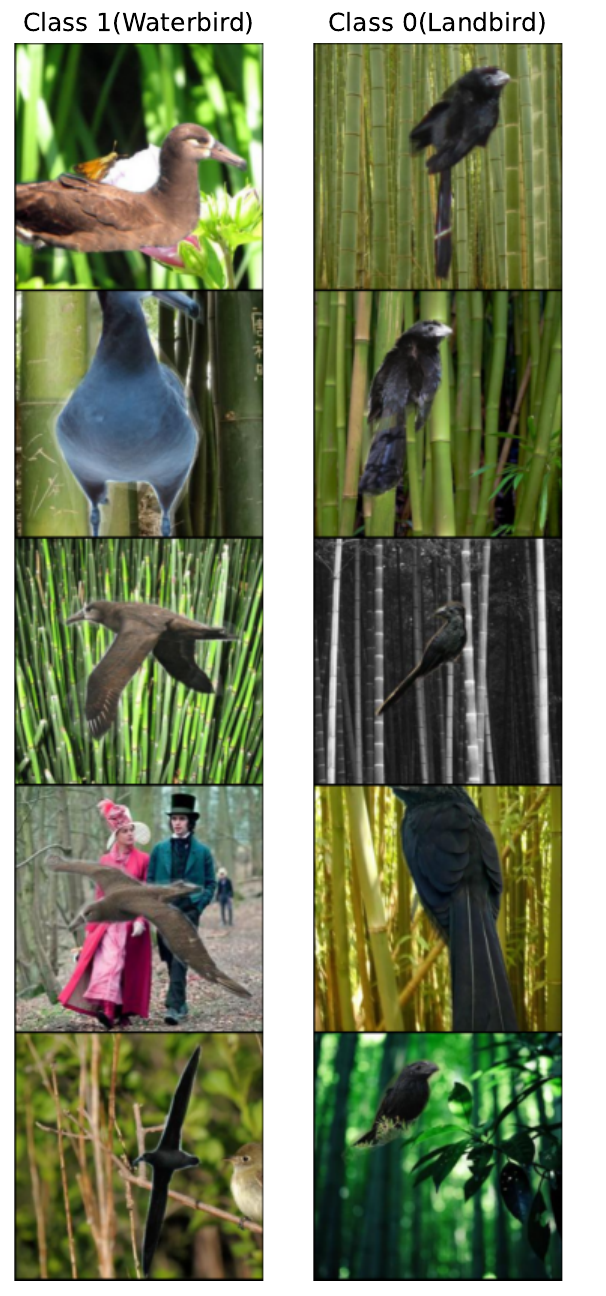}
\end{subfigure}
\begin{subfigure}
  \centering
  \includegraphics[scale=0.37]{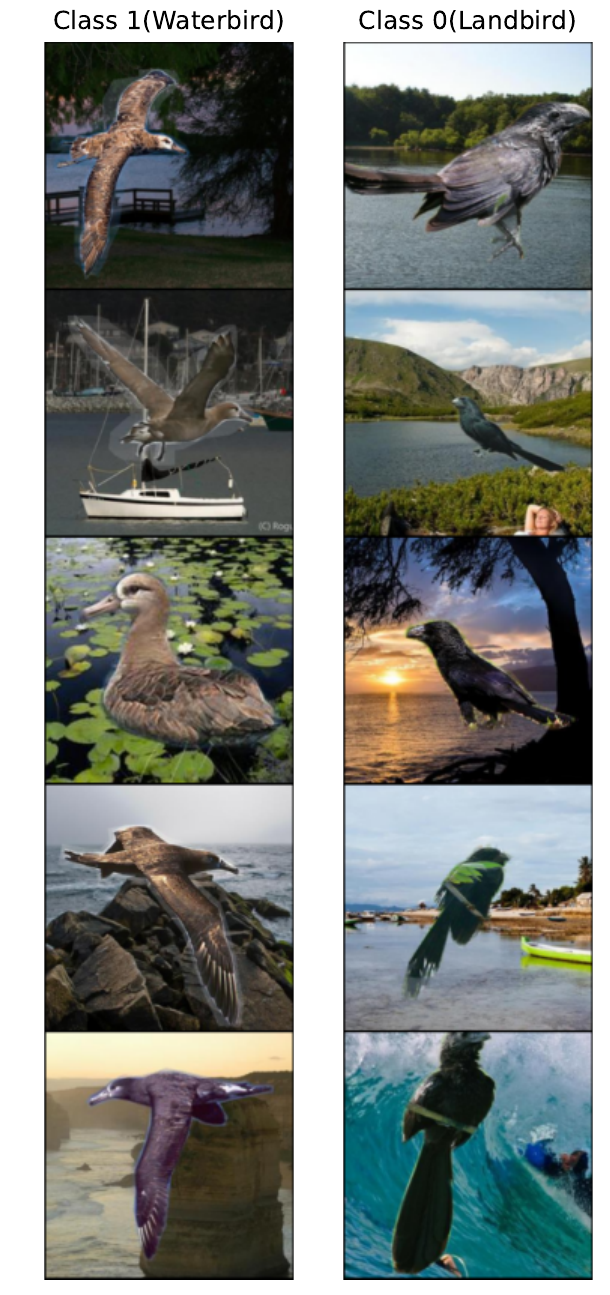}
\end{subfigure}
   \vspace{-2mm}
\caption{Images from Waterbirds dataset with different discovered concepts through our method. Here we arbitrarily select two of these concepts,  which can be interpreted as trees/bamboo background (left) and water background (right), and show input samples from each of these.}
\label{fig:clusters}
\vspace{-3mm}
\end{figure}

In \textbf{Stage 1},  we propose to vector quantize  semantic grouping representations 
    into discrete concepts (\cref{subsec:learning_dictionary}), enabling  the association of each input with relevant sets of concepts (see Fig \ref{fig:clusters}) and facilitating the calculation of concept occurrence statistics across the dataset. 
     
    In \textbf{Stage 2}, we utilize the occurrence statistics of concepts via importance sampling to train a separate classifier (\cref{subsec:concept_sampling}). The architecture of the classifier is inconsequential; the key contribution lies in the concept-aware sampling procedure, bridging object-centric representation learning and learning under subpopulation shifts.
    
    Integrating Stages 1 and 2 introduces {CoBalT} (Concept  Balancing Technique) tailored for robust classification.  We evaluate CoBalT across the CMNIST, Waterbirds, CelebA, Urban Cars and ImageNet-9 (IN-9L) datasets,  demonstrating improvements without the need for group annotations  (\cref{sec:experiments}). We achieve a 3\% improvement on Waterbirds compared to state-of-the-art (SOTA) group agnostic methods like MaskTune \citep{asgari2022masktune}, ULA \citep{Tsirigotis2023GroupRC} and XRM \citep{Pezeshki2023DiscoveringEW}, remain competitive on CelebA, and achieve 1--2\% improvement on challenging  IN-9L test sets while maintaining original test set performance. We also show $3\%$ improvement over SOTA baseline requiring group annotation in the Urban Cars dataset containing multiple spurious correlations per class.

\section{Related Works}
\label{sec:related_works}

\textbf{Robust training}: Our approach extends existing methodologies for robust classification model training, particularly addressing the costly acquisition of group labels in real-world data. Unlike GDRO \citep{Sagawa2020Distributionally}, which optimizes for the worst group-level error, and its semi-supervised extension, SSA \citep{nam2022spread}, our method is tailored for scenarios lacking sufficient labeled group data. Additionally, methods like DFR \citep{kirichenko2022last} and AFR \citep{Qiu2023SimpleAF} retrain the classification layer with group-balanced datasets and ensure feature reweighing, requiring group-labeled training/validation data, a necessity we circumvent. ULA \citep{Tsirigotis2023GroupRC} employs a Self-Supervised Learning (SSL) pre-trained model's predictions as a bias proxy, while MaskTune \citep{asgari2022masktune} assumes predictions from Empirical Risk Minimization (ERM) models to be biased. To train an unbiased model, the former adjusts the classifier's logits during debiasing training, and the latter masks out the input data based on the saliency map of the prediction.

\textbf{Group inference methods}: Obtaining group labels in real-world data is often costly. Several methods have been proposed for inferring group labels initially, followed by robust model training. LfF \citep{nam2020learning} uses two models, where the second model is trained using examples with higher loss in the first model. This approach contrasts with GEORGE \citep{sohoni2020no}, which clusters representations from the first stage ERM model to infer group information and then trains a second model using GDRO. Similarly, JTT \citep{liu2021just} and and CNC \citep{zhang2022correct}identify minority groups based on miss-classifications of the first stage ERM model; however, JTT continues with ERM to train the robust model, while CNC uses contrastive learning to align representations of minority examples with the majority. These methods either rely on extra group annotation or fail in the presence of multiple unbalanced minority groups and noisy examples \citep{yang2024identifying}.

SPARE \citep{yang2024identifying} separates spurious information in the early stages of training and uses k-means clustering to differentiate between minority and majority groups but relies on validation group annotation data to determine the specific epoch for separation. Conversely, our approach does not depend on group-annotated data for epoch identification; instead, we utilize a self-supervised method combined with spatial decomposition to separate spurious and non-spurious information effectively.The recently introduced XRM \citep{Pezeshki2023DiscoveringEW}  identifies groups within training and validation datasets through model prediction errors, operating under the assumption that models inherently learn spurious correlations. This methodology could be detrimental in scenarios where such an assumption does not hold true \citep{yong2022zin}.

{\bf Concept discovery.}
To address spurious correlations, DISC \citep{wu2023discover} introduces a human-interpretable concepts bank and \citet{moayeri2023spuriosity} rank data points by the spurious concepts that they contain. However, they require additional annotations of potential spurious features, posing practical challenges and limiting its general applicability. Learning abstract representations from images by decomposing them into higher-level concepts without human annotations has been explored in previous work. A recent development is slot attention \citep{locatello2020object}, which groups spatially repetitive visual features by imposing an attention bottleneck. This method and its variants have been successfully applied to discover object-centric concepts in synthetic datasets \citep{locatello2020object, Engelcke2021GENESISV2IU, zhang2023unlocking}. However, they face challenges when applied to complex real-world data. \citet{Seitzer2022BridgingTG} hypothesized that reconstructing the pixel space as a learning objective might not introduce enough inductive bias to facilitate the emergence of objects or concepts in real data. As a solution, they propose reconstructing the features from the self-supervised pre-trained DINO model \citep{Caron2021EmergingPI}. With similar motivation, \citet{wen2022self} employs a joint embedding teacher-student architecture, similar to \citet{Caron2021EmergingPI}, where the student model attempts to predict the concept representations of the teacher network. We extend this work to discover discrete symbols-like concepts by applying vector quantization \citep{Oord2017NeuralDR} to continuous concept representations aiding compositional reasoning of images, such as identifying common groups or attributes in the dataset like humans.

\begin{figure*}[t]
   \centering
   \includegraphics[width=.65\textwidth]{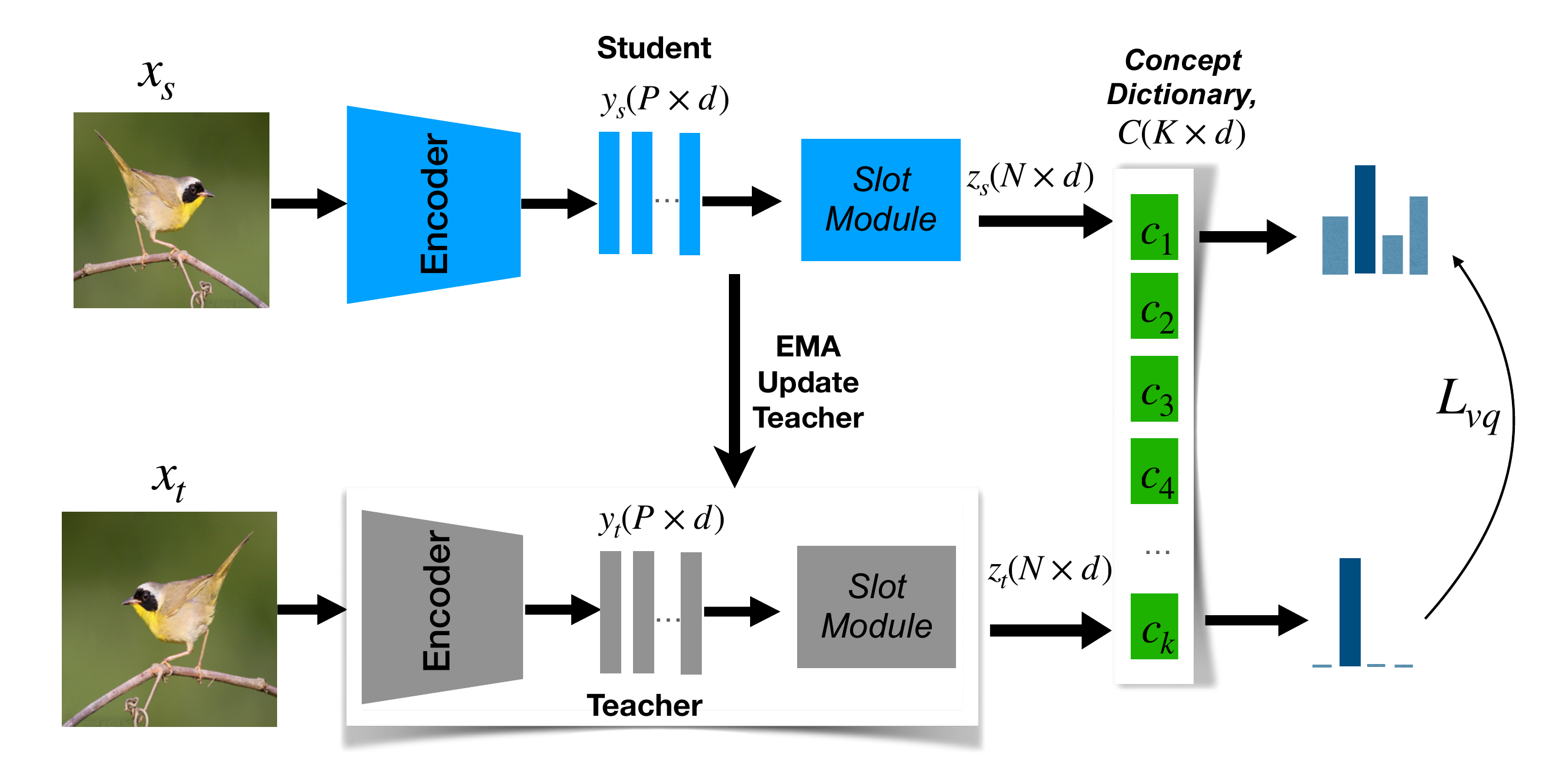}
   \caption{Architecture for learning slots and clustering without human annotation. $x_s$ and $x_t$ are two different augmented views of the same image. The teacher and student encoders project the augmented images into different patches $y_t$ and $y_s$ respectively, which are subsequently decomposed into concept representations $z_t$ and $z_s$ by slot attention. Then $z_t$ and $z_s$ are clustered into different concepts of dictionary $C$ using a vector-quantization. Stop-gradient is applied to teacher branch and the teacher encoder and slot module parameters are updated through the exponential moving average of the student encoder and slot module parameters.}
\label{fig:concept_learining_arch}
\vspace{-2mm}
\end{figure*}

\section{Method}
\label{sec:methodology}

Unlike existing methods, our goal is \emph{not} to discover the subgroups of a dataset specifically, but more general \emph{concepts}.
For example, while groups in the Waterbirds dataset are explicitly  defined to be the product of classes with some binary background attribute, $\{\text{water bird}, \text{land bird}\} \times \{\text{water background}, \text{land background}\}$,  the concepts could capture dataset-independent ideas such as blue bird, street background, or short beak.

We base our approach on the two-stage training procedure common in the literature, with the first stage determining some information about the training data and the second stage using this information to perform robust training.

The first stage combines two key components:
\begin{enumerate}
    \item Spatial clustering \citep{Caron2021EmergingPI, wen2022self}, which groups pixels into semantic regions (\cref{subsec:overview_existing}).
    While our approach in this paper is based on the method by \citet{wen2022self}, in principle, the requirement is simply for an unsupervised representation learner that decomposes the input into objects.
    
    \item A novel vector clustering technique we call \emph{concept dictionary learning}  (\cref{subsec:learning_dictionary}), achieved through vector quantization \citep{Oord2017NeuralDR}. This process discretizes the information of the slots into distinct concepts, which are more manageable compared to continuous representations of semantic regions. For example, instead of storing details about the specific shape and appearance of a bird, this clustering identifies broader concepts like bird types, which offer greater utility across various inputs. These concepts encompass typical foreground objects such as cats and dogs, background elements like land and sky (see Fig. \ref{fig:clusters}), or other abstract notions not as readily interpretable as individual words.
\end{enumerate}
\vspace{-2mm}

The key aspect of our proposed second component is its independence from human labeling, achieved through leveraging the self-supervised learning setup of the first component.
This lack of reliance to human labeling offers significant advantages, particularly in complex datasets.
For example, when dealing with large datasets like ImageNet, determining relevant subgroups across the images is challenging due to the vast number of possibilities.
Spurious correlations in a dataset are likely to vary depending on the specific task being performed with the dataset. Without pre-labeling every conceivable group (which is clearly infeasible), identifying the subgroups necessary to address spurious correlations seems nearly impossible.

By adopting a data-driven approach where concepts are learned, we can discover concepts that a model inherently relies on. 
However, this approach has the potential drawback of weakening the connection between a learned concept and a concept that humans readily understand. One advantage of an object-centric decomposition, as demonstrated  by methods like \citet{wen2022self}, is that the spatial grouping of a concept provides humans with additional insight into that concept represents.

\vspace{-1mm}
\subsection{Architecture} 
\label{subsec:overview_existing}
The model architecture used for concept learning shares the same
overall structure as many recent self-supervised approaches to representation learning~\citep{Caron2021EmergingPI, Grill2020BootstrapYO, zbontar2021barlow, chen2020simple}.  Following  \citet{wen2022self}, we employ  a two-branch network where the branches are structurally similar but asymmetric in parameter weights.
Each branch comprises an \emph{encoder} that outputs patch representation vectors of the input image, a \emph{projector} that transforms the representations into an embedding space, and a \emph{slot module} where spatial patch representation vectors are semantically grouped into concept representations. 
Our focus lies on building our model based on the output of the slot module. The overall architecture 
is illustrated and briefly described in \cref{fig:concept_learining_arch}, with detailed information provided in \autoref{app:detailed_arch}. 

More precisely, we will utilize the slots of the student and teacher branches, $z_s \in R^{N \times d}$ and $z_t \in R^{N \times d}$, where the hyperparameters $N$ represents the number of slots and $d$ denotes the dimensionality of each slot.
Each slot serves as a semantic grouping of an area in the input; for instance, a slot could correspond to a single object in the image.

\subsection{Concept Dictionary}
\label{subsec:learning_dictionary}
For the next step, we aim to discover meaningful discrete concepts from these spatially-decomposed semantic groupings.
To do so, we employ vector quantization \citep{Oord2017NeuralDR} which acts as a learned discretization or clustering mechanism that effectively clusters similar concepts in the training data into distinct categorical concepts. 

This is done by utilizing a codebook that we call \textbf{concept dictionary} $C \in R^{K\times d}$ with $K$ vectors of dimension $d$, each of which corresponds to a symbolic concept (e.g. water, tree, bird, etc.) that we want to learn. Note that we do not supervise these concepts in any way -- these words simply denote possible meanings we could assign to these concepts post-hoc. We assign each slot (vector representation) to a discrete symbolic concept by learning a categorical distribution over the entries in the dictionary.

Given a randomly initialized concept dictionary $C$, we associate each slot (for student and teacher branches) to a concept in the concept dictionary through distributions $p_s \in \mathbb{R}^{K \times N}, p_t \in \mathbb{R}^{N}$ by seeking the most similar concept:
\begin{align}
(p_s)_{ij} & = \frac{\exp(-\|C_i - (\bar{z}_{s})_j\|^2_2/\tau_s)}{\sum_{t=1}^K \exp(-\|C_t - (\bar{z}_{s})_j\|^2_2/\tau_s)} \\
(p_t)_{i} & = \argmax_j -\|C_j - (\bar{z}_{t})_i\|^2_2
\label{eq:clustering}
\end{align}
where $C_i, (\bar{z}_{s})_j \in \mathbb{R}^d$, $\bar{z}_{s} = z_{s} / ||z_{s}||$, $\bar{z}_{t} = z_{t} / ||z_{t}||$ (the $i$-th slot of $z_s$ and $z_t$ are normalized to have unit L2 norm), and $\tau_s$ is a temperature hyperparameter. 
For the teacher branch, inspired by \citet{Caron2021EmergingPI}, rather than taking a $\softmax$, we use a sharpened distribution.
In particular, we employ the $\argmax$ to facilitate a hard assignment into a one-hot representation. This hard assignment ensures that each slot is associated with a single distinct concept. We then use these as the supervision signal for the student branch, where we encourage each slot representation of the student to also be assigned to a single concept vector. This difference in $\softmax$ and $\argmax$ has the benefit of making the distributions of the student and teacher branches different, which avoids the representation collapse problem mentioned in \citet{Caron2021EmergingPI}.


Following \citet{roy2018towards}, throughout training at each step, $C$ is updated by the exponential moving average of batch-wise teacher concept representations $z_t$ as follows:
\begin{align}
    C_j = \alpha_c \cdot C_j + (1 - \alpha_c) \cdot \sum_i \mathbbm{1}\{(p_{t})_{i}=j\} (z_t)_i
\end{align}
with $\alpha_c$ is the update rate of the codebook. We set it to $0.9$ for all our experiments.

\paragraph{Loss}

As for the learning objectives, in addition to the losses proposed by \citet{wen2022self} (see Appendix~\ref{eq:indicator}), we include a novel term ${\mathcal{L}_{vq}}$, motivated as follows.  Since we do not have any explicit human supervision of concepts, we exploit the assignment of concepts of the teacher to supervise the student. The purpose of the loss term \(\mathcal{L}_{vq}\) is to ensure the consistency of the prediction between the slot representations of the teacher and the student. We encourage this alignment by distilling the teacher's prediction of discrete concepts to the student with a cross-entropy loss, which is calculated as follows:
\begin{equation}
    {\color{red!50!black} \boldsymbol{\mathcal{L}_{vq}}} = - \sum_{i=1}^N \sum_{j=1}^K \mathbf{I}(i) \mathbbm{1}\{(p_{t})_{i}=j\} \log (p_{s})_{ij}
\end{equation}
where $\mathbf{I(i)}$ is the indicator function that avoids calculating the loss for the slot where the student and teacher does not have any common patch assignment. 
Details of how this is calculated are described in Appendix~\ref{eq:indicator}. 

We then include this objective as a term in the overall loss of \citet{wen2022self}:
\begin{equation}
\boldsymbol{\mathcal{L} = {\mathcal{L}_{dis}} +{\mathcal{L}_{con}} + {\color{red!50!black} \mathcal{L}_{vq}}}
\label{eq:final_loss}
\end{equation}
where ${\mathcal{L}_{dis}}$ governs {\it attention distillation} from teacher to student  and $\mathcal{L}_{con}$ is a {\it contrastive loss} between slot representations to avoid redundancy and encourage diversity. These losses   are 
described in detail in~\cref{app:loss_wen_el}.

This concludes the first stage of our training process. To recap, we extract slot representations following the methodology in \citet{wen2022self}, then compute concept distributions $p_s$ and $p_t$ over the concept dictionary $C$, which is incrementally updated based on assignments from the teacher branch. Our learning objective is designed to distill the teacher concept distribution to the student. Through this process, we establish the association of training samples with sets of concepts. This information will be utilized in the subsequent section.

\subsection{Training a Robust Classifier} 
\label{subsec:concept_sampling}

In the second stage, we train a separate classifier based on the concepts learned in the first stage, which are considered fixed. Integrating this information into the training process offers various possibilities. Our approach draws inspiration from previous works \citep{Sagawa2020Distributionally, yang2024identifying} where, if ground-truth subgroups are known, adjusting the subgroup sampling rate evenly is the most effective method. We adapt this concept to our framework, modifying it to suit learned concepts rather than ground-truth subgroups. However, this adaptation presents challenges, such as each data point belonging to multiple concepts instead of a single subgroup, and the occurrence of each concept in multiple classes at varying frequencies.

\paragraph{Sampling method}

Our core approach involves adjusting the sampling rate of samples to ensure an even representation of concepts and, when feasible, an even representation of classes within those concepts. This entails sampling prevalent concepts less frequently and rare concepts more frequently. Additionally, within each concept, we aim to maintain balanced representation of labeled classes. By doing so, we bias the classifier training towards rarer concepts while striving to balance classes within a concept whenever possible.

This strategy is guided by the understanding that minority groups, characterized by rarer concepts within a class, are more susceptible to misclassification due to concept overlap. Notably, our sampling method differs from the weighting scheme proposed by \citep{yang2024identifying}, which contrasts between groups within the same class. Instead, our approach focuses on contrasting between samples from the same concept but belonging to different classes.

Within a cluster $c$, we have multiple classes with $T_{c, y}$ represents the samples from the cluster $c$ and class $y$. We compute the weight and probability of sampling that class within the cluster as:
\begin{equation} 
w_{c, y}=\frac{1}{|T_{c, y}|}, \quad p_{c, y} =\frac{w_{c, y}^{\lambda}}{\sum_{\hat{y}} w_{c, \hat{y}}^{\lambda}}
\label{eq:sampling_p}
\end{equation}
where $\lambda$ is a sampling factor, a hyperparameter.
\citet{yang2024identifying} recommend to increase $\lambda$ from the default of 1 when the inter-concept groups are not well separable.
The choice of this hyperparameter can be guided by the average silhouette score \citep{rousseeuw1987silhouettes}, which measures how well the clusters are separated. In our case, it reflects the degree of distinction between groups from one cluster to the groups of the other cluster.

\begin{algorithm}
\caption{Batch Sampling Strategy}
\label{alg:batch-sampling}
$K$: clusters with samples of different classes \\ $n$: batch size \\ $T_{c, y}$: Set of samples belonging to cluster $c$ and class $y$ \\
\begin{algorithmic}
    \STATE Initialize $batch \leftarrow \{\}$
    \FOR{$i = 1$ to $n$}
        \STATE $c \leftarrow$ uniformly select a cluster from $1$ to $K$
            \STATE $w_{c, y} \leftarrow$ calculate weights $\frac{1}{|T_{c, y}|}$
            \STATE $y \leftarrow$ select a class with $p_{c, y} = \frac{w_{c, y}^{\lambda}}{\sum_{\hat{y}} w_{c, \hat{y}}^{\lambda}}$
        \STATE $b \leftarrow$ select a sample from $c$ of class $y$
        \STATE $batch \leftarrow batch \cup \{b\}$
    \ENDFOR
    \STATE \textbf{return} $batch$
\end{algorithmic}
\end{algorithm}

\vspace{-3mm}
\subsection{Early stopping}

As demonstrated by \citet{YoubiIdrissi2021SimpleDB}, having access to group information is crucial for effective model selection, particularly in scenarios involving spurious correlations. In our experiments, we explore three distinct model selection strategies by altering the criteria for early stopping:

\begin{enumerate} 
\item \SLOT{}$_{hg}$: This strategy relies on human-annotated worst-group labels.
\item \SLOT{}$_{ig}$: Here, we utilize the inferred worst group.
\item \SLOT{}$_{avg}$: This strategy employs the average validation accuracy as the criterion for early stopping.
\end{enumerate}

While \SLOT{}$_{hg}$ offers the advantage of leveraging human annotations, it also reintroduces dependency on manual labeling. Consequently, we generally prefer settings where \SLOT{}$_{ig}$ and \SLOT{}$_{avg}$ are more suitable.

For \SLOT{}$_{ig}$, our approach involves inferring groups from the discovered concepts. Each group is defined by the unique combination of class and concept. For instance, if we have two concepts and three classes, we would generate six groups accordingly. It is important to note that these inferred groups may not align with the ground-truth groups in the dataset, if such labels are even available. Nevertheless, our method utilizes these inferred groups as an early stopping criterion.
\section{Experiments}
\label{sec:experiments}
To illustrate the effectiveness of our spatial concept discovery and sampling strategy, we investigate two challenging scenarios where training a robust classifier using empirical risk minimization (ERM) with i.i.d. (independent and identically distributed) sampling faces significant difficulties.

\textbf{Scenario 1:} Classification complicated by class imbalance and attribute imbalance with single spurious correlation. In this scenario, markedly underrepresented, and attributes within classes exhibit uneven distributions in the training data. This presents considerable challenges for an ERM-trained model, particularly concerning under-represented attributes. 

\textbf{Scenario 2:} When data containing two spurious correlations per class, \citet{li2023whac} shows the limitations of SOTA methods that present 'whack-a-mole' behavior: mitigating one spurious correlation but amplifying the other.

\textbf{Scenario 3:} Test data has attributes not present in the training data, requiring attribute generalization. For example, if the training set has cows on grassland and rarely on a beach, the test set might have cows on a volcano. This scenario is demanding as the classifier must recognize and generalize unknown attributes. Moreover, merely defining subgroups in this scenario is inherently challenging.

\subsection{Datasets}

\begin{figure*}[h]
   \centering
   \includegraphics[width=.80\textwidth]{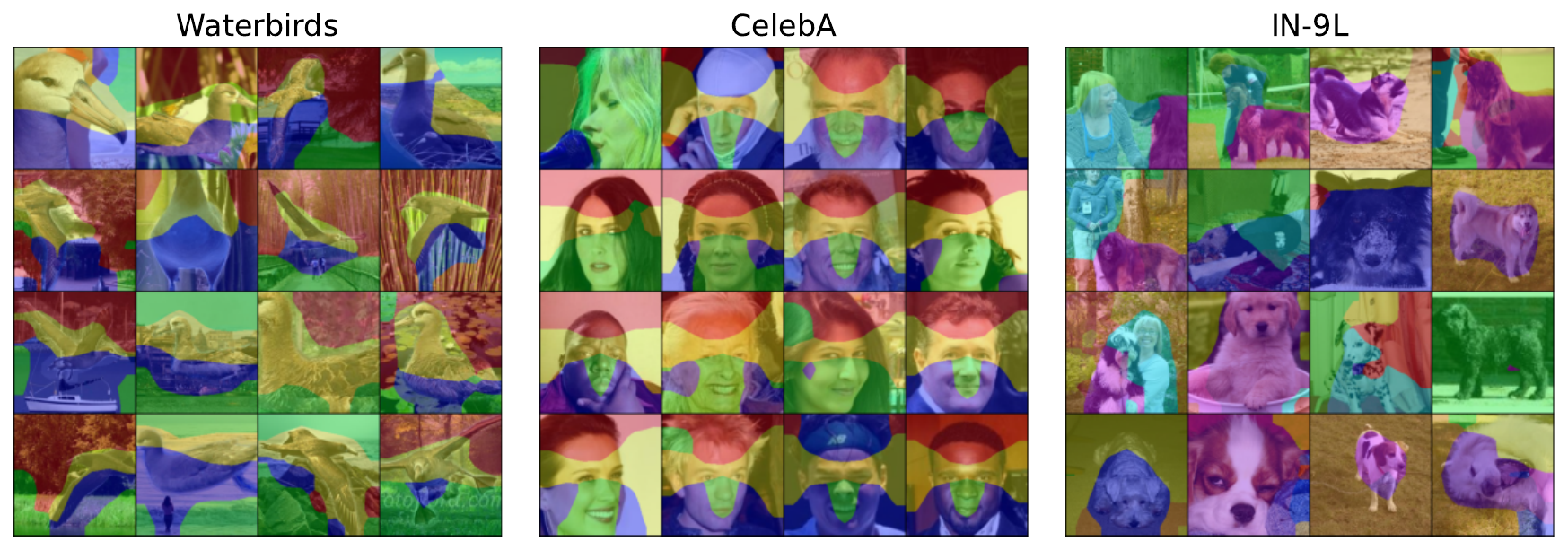}
   \vspace{-3mm}
   \caption{Segmented regions with slots in different datasets. $\#$ of slots used for Waterbirds, CelebA and ImageNet-9 is $4$. The pixels in the images are grouped by slots and represent high-level concepts such as body parts of birds and backgrounds like trees, water, and so on in Waterbirds; humans, animals, grasses, and so on in \emph{IN-9L} and nose, head, and so on in CelebA.}
   
   \label{fig:slot_attention_segmentation}
\vspace{-3mm}
\end{figure*}

Considering the scenarios outlined, we train our model using the following publicly available datasets: CMNIST \citep{alain2015variance}, CelebA \cite{Liu2014DeepLF}, Waterbirds \cite{Sagawa2020Distributionally}, UrbanCars \citep{li2023whac}, Background Challenge ImageNet-9 \cite{xiao2020noise}.

\begin{table*}[ht]
\centering
\caption{CMNIST results with LeNet-5, and Waterbirds and CelebA results with ImageNet pre-trained ResNet50. Other model results are reported from \citet{yang2024identifying, asgari2022masktune, Pezeshki2023DiscoveringEW, Tsirigotis2023GroupRC}.
The upper section uses human annotated group information for training and validation, the middle for validation only, and the bottom does not use group annotation. Best results are highlighted in each section. 
\SLOT{}$_{hg}$, \SLOT{}$_{ig}$, \SLOT{}$_{avg}$ are our trained models where early stopping is done by human-annotated worst group, inferred worst group, and average validation accuracy respectively.}
\vspace{0.25cm}
\resizebox{0.8\linewidth}{!}{
\begin{tabular}{c|cc|cc|cc|cc}
\toprule
& \multicolumn{2}{|c|}{Group Label} & \multicolumn{2}{|c|}{CMNIST} & \multicolumn{2}{|c|}{Waterbirds} & \multicolumn{2}{c}{CelebA} \\
\hline
Method & \textbf{Train} & \textbf{Val} & Worst Group & Average & Worst Group & Average & Worst Group & Average \\
\hline
GB & \checkmark & \checkmark & \textbf{82.2±1.0} & \textbf{91.7±0.6} & 86.3±0.3 & 93.0±1.5 & 85.0±1.1 & 92.7±0.1 \\
DFR$^{Tr}$~\citep{kirichenko2022last} & \checkmark & \checkmark & - & - &  90.4±1.5 & \textbf{94.1±0.5} & 80.1±1.1 & 89.7±0.4 \\
DFR$^{Val}$~\citep{kirichenko2022last} & \checkmark & \checkmark & - & - &  \textbf{91.8±2.6} & 93.5±1.4 & 87.3±1.0 & 90.2±0.8 \\
GDRO \citep{Sagawa2020Distributionally} & \checkmark & \checkmark &  78.5±4.5 &	90.6±0.1 & 89.9±0.6 & 92.0±0.6 & \textbf{88.9±1.3} & \textbf{93.9±0.1} \\
DISC \citep{wu2023discover} & \checkmark & \checkmark &  - & - & 88.7±0.4 & 93.8±0.7 & - & -\\ 
\hline
GEORGE \citep{sohoni2020no} & $\times$ & \checkmark & 76.4±2.3 & 89.5±0.3 & 76.2±2.0 & \textbf{95.7±0.5} & 54.9±1.9 & \textbf{94.6±0.2} \\
LfF \citep{nam2020learning} & $\times$ & \checkmark & 0.0±0.0 &	25.0±0.0 & 78.0 & 91.2 & 77.2 & 85.1 \\
CIM \citep{taghanaki2021robust} & $\times$ & \checkmark & 0.0±0.0 & 36.8±1.3 & 77.2 & 95.6 & 83.6 & 90.6 \\
JTT \citep{liu2021just} & $\times$ & \checkmark & 74.5±2.4 &	90.2±0.8 & 83.8±1.2 & 89.3±0.7 & 81.5±1.7 & 88.1±0.3 \\
CnC \citep{zhang2022correct} &  $\times$ & \checkmark & 77.4±3.0 & 90.9±0.6 & 88.5±0.3& 90.9±0.1 & 88.8±0.9 & 89.9±0.5 \\
SPARE~\cite{yang2024identifying} &  $\times$ & \checkmark & \textbf{83.0±1.7}  &  91.8±0.7 & 89.8±0.6 & 94.2±1.6 & \textbf{90.3±0.3} & 91.1±0.1 \\
AFR \citep{Qiu2023SimpleAF} &  $\times$ & \checkmark & - & - & 90.4±1.1 & 94.2±1.2 & 82.0±0.5 & 91.3±0.3 \\
\SLOT{}$_{hg}$ (ours)  & $\times$ & \checkmark & 79.0±4.3 & \textbf{96.6±1.8} & \textbf{90.6±0.7} & 93.7±0.6 & 88.0±2.5 & 92.3±0.7 \\
\hline
ERM & $\times$ & $\times$ & 0.0±0.0	& 20.1±0.2 & 62.6±0.3 & \textbf{97.3±1.0} & 47.7±2.1 & \textbf{94.9±0.3} \\
MaskTune \citep{asgari2022masktune} & $\times$ & $\times$ & - & - & 86.4±1.9 & 93.0±0.7 & 78.0±1.2 & 91.3±0.1 \\
ULA \citep{Tsirigotis2023GroupRC} & $\times$ & $\times$ & \textbf{75.1±0.8} & - & 86.1±1.5 & 91.5±0.7  & 86.5±3.7 & 93.9±0.2  \\
XRM \citep{Pezeshki2023DiscoveringEW} & $\times$ & $\times$ & 70.5 & - & 86.1 & 90.6 & \textbf{89.8} & 91.8 \\
\SLOT{}$_{ig}$ (ours) & $\times$ & $\times$ & 73.5±2.1 & 96.0±1.6 & 89.0±1.6 & 92.5±1.7 & 89.2±1.2 & 92.3±0.6  \\
\SLOT{}$_{avg}$ (ours)  &  $\times$  & $\times$  & 74.5±2.0 &  \textbf{96.2±2.0} & \textbf{90.6±0.7} & 93.8±0.8 & 81.1±2.7 & 92.8±0.9 \\
\bottomrule
\end{tabular}
}
\label{tab:wb_ca_results}
\vspace{-2mm}
\end{table*}

\textbf{Scenario 1. }
The \textbf{CMNIST} dataset \citep{alain2015variance} contains colored versions of the MNIST digits \citep{lecun1998gradient}. We use the challenging 5-class setup from \citep{zhang2022correct}, where each class pairs two digits, with 99.5\% of training samples in each class spuriously correlated to a unique color.

The \textbf{CelebA} dataset \citep{liu2015deep} shows a significant class imbalance in gender (male/female) and hair color (dark/blonde). Most of the male images (162,770) have dark hair, while only 1,387 (0.85\%) have blonde hair. This imbalance risks bias, potentially causing the model to associate gender with hair color.

The \textbf{Waterbirds} dataset, as detailed in \citep{Sagawa2020Distributionally}, has two classes: landbirds and waterbirds. The background: land or water acts as a spurious attribute. The common instances (waterbird, water) and (landbird, land) make it challenging to differentiate the bird type from the spuriously correlated background.

\textbf{Scenario 2. }
The \textbf{UrbanCars} dataset \citep{li2023whac} includes two classes (urban and country vehicles) along with two incidental spurious attributes: (1) background (BG): city vs. countryside and (2) co-occurring objects (CoObj): fireplug and stop sign vs. cows and horses.

\textbf{Scenario 3. }
We utilize the \textbf{Background Challenge ImageNet-9 (IN-9L)} dataset \citep{xiao2020noise}, derived from a subset of ImageNet known as ImageNet-9. This dataset is purposefully crafted to assess the robustness of models against background variations. It encompasses four distinct types of background modifications in its test sets: 

\begin{itemize} 
\item \textbf{Original:} Maintains the original background. 
\item \textbf{Mixed-same:} Replaces the background taken from another image within the same class. 
\item \textbf{Mixed-rand:} Replaces the background taken from a randomly selected image.
\item\textbf{Only-FG:} Eliminates the background entirely, leaving only the foreground object. 
\end{itemize}

This dataset challenges classifiers to remain robust to background changes, serving as a benchmark for evaluating a model's ability to generalize and focus on primary object features despite background variability or absence.

\subsection{Results} 
We present additional results and ablations in \autoref{app:addional_result}.

\subsubsection{Scenario 1 (CMNIST, Waterbirds, CelebA)}

As shown in \cref{tab:wb_ca_results}, our evaluation on Waterbirds, CelebA and CMNIST showcases the effectiveness of our approach, which achieves superior or comparable performance compared to methods that do not rely on human-annotated group labels. Particularly noteworthy is \SLOT{}$_{ig}$,  which outperforms  in worst-group accuracy the recent methods ULA \citep{Tsirigotis2023GroupRC} and XRM \citep{Pezeshki2023DiscoveringEW} by nearly $3\%$ on Waterbirds, while also demonstrating competitive performance on CelebA and CMNIST with an average accuracy similar to the other methods. 

Even when selecting the model based on the average validation accuracy (\SLOT{}$_{avg}$),  without attempting to infer groups, our model still demonstrates competitive results. Unlike other baselines that leverage human-annotated group-labeled training or validation sets for early stopping or hyperparameter tuning (as detailed in \autoref{app:addional_result}), 
our method makes group inferences for both training and validation data without relying on human labels.

\begin{figure}[htb]
   \centering
   \includegraphics[width=0.7\linewidth]{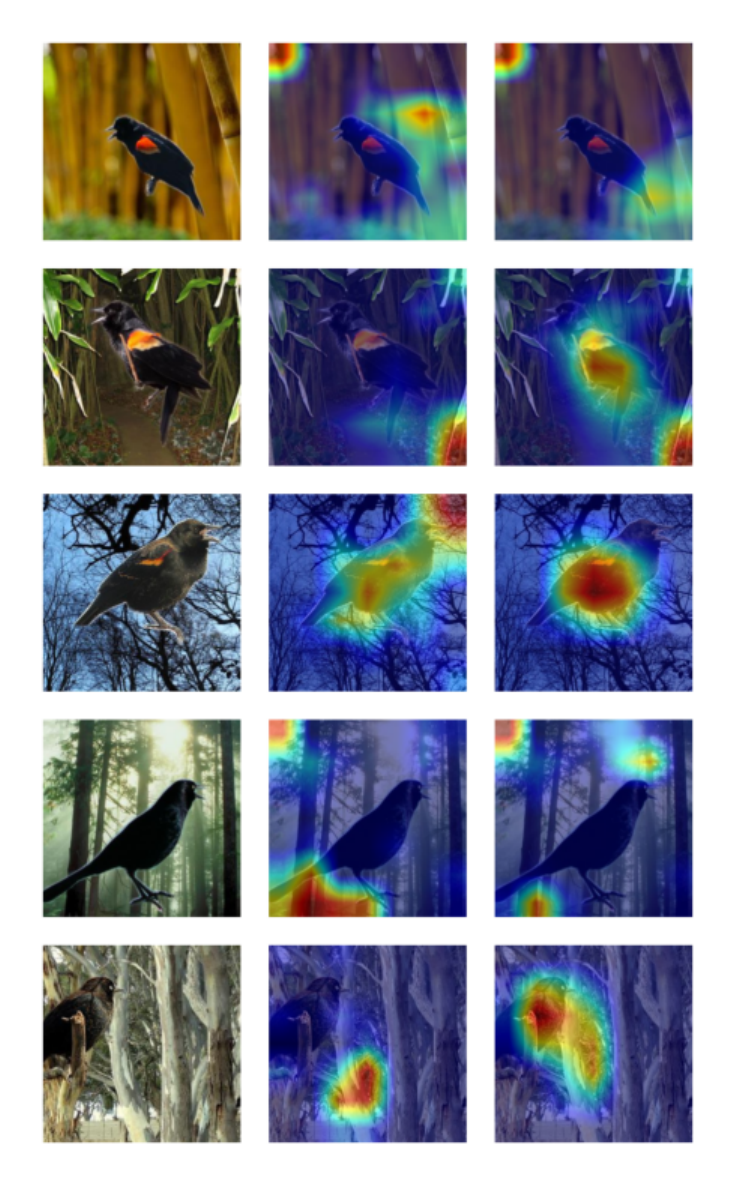}
   \vspace{-2mm}
   \caption{Gradcam heat-map on Waterbirds dataset (from left to right Input, ERM and \SLOT{}  respectively in three columns). ERM models spuriously correlates to background information for classifying bird types whereas our methods reduce the spurious correlation by focusing on image regions that contain birds.}
   \label{fig:waterbirds_gradcam}
\vspace{-5mm}
\end{figure}

Furthermore, our method exhibits similar performance to other methods employing group annotations. We provide visualizations of the feature attributions of ERM and our method in the Waterbirds dataset, as illustrated in \cref{fig:waterbirds_gradcam}, demonstrating that our method relies less on spurious backgrounds compared to ERM.

\subsubsection{Scenario 2 (Urban Cars)}
When multiple spurious correlations are present in the datasets, existing methods show a Whack-a-mole behavior, where mitigating one spurious correlation amplify the other in minority groups. \emph{CoBalT} is good at identifying abstractions from data, which help to identify multiple spurious attributes and thus improve spurious correlation.

From Table ~\ref{tab:urban_cars}, we can see that when there are two spurious correlations such as backgrounds and co-occurring objects, our method overcomes these correlations and achieves SOTA performance outperforming \emph{SPARE} by $3\%$. We also see that methods like \emph{EIIL}, \emph{LfF} and \emph{JTT} struggle to perform better in the presence of multiple spurious correlations.

\begin{table}[t]
\centering
\caption{Results from the Urban Cars dataset using ResNet-50, when there are multiple spurious correlations (BG+CoObj) exist.}
\vspace{0.25cm}
\resizebox{0.65\linewidth}{!}{
\begin{tabular}{c|cc}
\toprule
 Method & Worst Group & Average \\
\hline
ERM & 28.4 & \textbf{97.6} \\
EIIL & 	50.6 & 95.5 \\
GEORGE & 35.2 & 97.9 \\
LfF & 34.0 & 97.2 \\
JTT & 55.8 & 95.9 \\
SPARE & 76.9±1.8 & 96.6±0.5 \\
GDRO & 75.2 & 91.6 \\
\SLOT{}$_{ig}$ (ours)  & \textbf{80.0±2.8} & 96.3±0.6\\
\SLOT{}$_{avg}$ (ours)  & 76.8±6.5 & 97.3±0.7\\
\bottomrule
\end{tabular}
}
\label{tab:urban_cars}
\vspace{-1mm}
\end{table}

\subsubsection{Scenario 3 (ImageNet-9 Background)}
In the more realistic setting of the ImageNet-9 background challenge dataset, we assess the attribute generalization capability of our method. Training our model exclusively on the original ImageNet-9 trainset, without accessing the `mask-rand' subset where background images are randomly swapped, we select the model based on inferred worst group performance on the original validation set.

As illustrated in \cref{tab:bg_challenge_result}, our method (\SLOT{}$_{ig}$) outperforms MaskTune \citep{asgari2022masktune} by $1.1\%$ on Mixed-same, $1.5\%$ on Mixed-rand, and $1.9\%$ on Only-FG. Additionally, we observe improvements compared to other methods across all test sets. These results underscore the efficacy of our concept discovery method and the importance weight-based sampling strategy in learning task-relevant information and mitigating spurious correlations. Notably, our sampling technique for addressing imbalances within the training set remains effective even in scenarios where the imbalance is not readily apparent.

Many techniques used for Waterbirds and CelebA are inapplicable to this dataset due to the lack of inference groups. Our method, however, is more versatile and performs well across various scenarios.

\begin{table}[htbp]
\centering
\caption{Results on Background Challenge (ImageNet-9). Top rows based on ResNet-50 (ImageNet-Pretrained), 4 slots and codebook size 8. The results of other methods are taken from \citet{asgari2022masktune}.}
\vspace{0.25cm}
\resizebox{\linewidth}{!}{
\setlength{\tabcolsep}{1.5pt}
\begin{tabular}{c|cccc}
\toprule
Method & Original & Mixed-same  & Mixed-rand  & Only-FG \\
\hline
ERM  & \textbf{97.9} & 90.5 & 79.2 & 88.5 \\

CIM~\citep{taghanaki2021robust} & 97.7 & 89.8 & \textbf{81.1} & - \\
SIN~\cite{Sauer2021CounterfactualGN} &  89.2 & 73.1 & 63.7 & - \\
INSIN~\cite{Sauer2021CounterfactualGN} & 94.7 & 85.9 & 78.5 & - \\
INCGN~\cite{Sauer2021CounterfactualGN} & 94.2 & 83.4 & 80.1 & - \\
MaskTune~\citep{asgari2022masktune} & 95.6 & 91.1 & 78.6 & 88.1 \\
\SLOT{}$_{ig}$ (ours) & \textbf{97.9} & \textbf{91.2} & 80.1 & 90.0 \\
\SLOT{}$_{avg}$ (ours) & \textbf{97.9} & \textbf{91.2} & 80.3 & \textbf{90.1} \\
\bottomrule
\end{tabular}
}
\label{tab:bg_challenge_result}
\vspace{-3mm}
\end{table}

\subsubsection{Results without validation groups}
In our previous evaluations,  we selected the model by early stopping based on the worst group validation performance, with the groups being inferred on the validation data by our proposed method. To evaluate the impact of model selection, we now consider a scenario where we lack access to human-annotated validation groups for CelebA. In this case, other methods select the model based on average validation accuracy, as they typically rely on human-annotated validation groups.

From~\cref{tab:ca_without_val}, we can see that the performance of different methods substantially degrades when group-labeled validation data is unavailable for early stopping. Many of the group inference methods perform even worse than ERM, with the notable exception of MaskTune. However, MaskTune still performs significantly worse than our methods \SLOT{}$_{ig}$ and \SLOT{}$_{avg}$. This underscores the critical importance of having access to group-labeled data for many baseline methods to effectively work. 

In contrast, our method proves valuable by inferring groups in an unsupervised manner. when we perform early stopping based on average validation accuracy, akin to the baseline methods in this table, our method \SLOT{}$_{avg}$ significantly outperforms others, particularly on the worst group.

\vspace{-3mm}
\begin{table}[htbp]
\centering
\caption{Results from the CelebA dataset using ResNet-50 (when early stopping is not done using validation group labels for other methods). We do early stopping based on our inferred groups on the validation set without using validation group labels. The baseline results are taken from \citet{asgari2022masktune}.}
\vspace{0.20cm}
\resizebox{0.75\linewidth}{!}{
\begin{tabular}{c|cc}
\toprule
 Method & Worst Group & Average \\
\hline
ERM & 47.7±2.1 & \textbf{94.9±0.3} \\
CVaR DRO \cite{levy2020large} & 36.1 & 82.5\\
DivDis \citep{lee2023diversify} & 55.0 & 90.8 \\
LfF & 24.4 & 85.1 \\
JTT & 40.6 & 88.0 \\

MaskTune & 78.0±1.2 & 91.3±0.1 \\
\SLOT{}$_{ig}$ (ours)  & \textbf{89.2±1.2} & 92.3±0.6\\
\SLOT{}$_{avg}$ (ours)  & 81.1±2.7 & 92.8±0.9 \\

\bottomrule
\end{tabular}
}
\label{tab:ca_without_val}
\vspace{-1mm}
\end{table}

\subsubsection{Interpretation of concepts}
Our proposed method decomposes images into high-level concepts in an unsupervised way and clusters the images based on those concepts. Through the slot-based decomposition model, objectness or high-level concepts emerge in complex real-world data sets, which can be viewed through the attention map of each slot as in~\cref{fig:slot_attention_segmentation}. For example, in Waterbirds, the region grouped by slots belongs to parts of the body of birds and background such as trees, water, etc. In the IN-9L dataset, the slot distinguishes humans, animals, grass, etc. For CelebA, the model learns to separate the nose, eyes, and hair on the human face.

These decomposed slot representations are matched with a set of vector-quantized codes from the learned dictionary. Each code in the dictionary represents high-level abstract concepts. This approach effectively makes each code as the centroid of a specific cluster. By matching slot representations to the closest centroid, we can categorize an image into multiple distinct clusters. This allows us to identify and group images based on shared high-level concepts, despite the fact that they belong to different classes. Such an organization becomes particularly insightful when we observe images from varied classes clustering together. This clustering is based on the similarity of the underlying concepts these images represent. For example, images from different classes but with a common feature or concept might find themselves grouped in the same cluster (e.g. trees and water in~\cref{fig:clusters} respectively).

\subsubsection{Limitations without Spatially Separable Concepts}
Our method relies on the disentanglement of different concepts by clustering slots, which implicitly assumes spatial separability between different concepts. These concepts are assumed to be at the object level that might represent foreground/background spurious correlations. We hypothesize that because of this, on CMNIST and CelebA (Table \ref{tab:wb_ca_results}), where concepts are less spatially separable, our method's benefit is a bit limited compared to other datasets like Waterbirds, Urban Cars, Bar \citep{nam2020learning} (Table \ref{tab:bar}), where concepts have clear spatial regions. Further investigation in this direction is needed. In principle, one could try to learn a disentangled representation of the high level objects and use the disentangled factors of variations as concept. Work like \citet{Singh2022NeuralSB} that further factorizes the representation of each semantic region can be used to mitigate this issue by allowing one spatial region to be represented as a collection of concepts.

\subsubsection{Avoiding Representation Collapse}
Our concept discovery method is based on self-supervised Siamese representation learning, utilizing two parallel encoders: the student produces the source slot encoding and the teacher produces the target encoding. One of the main issues with this kind of encoder-only learning framework is representation collapse \cite{Hua2021OnFD}. During training, our method can obtain a degenerate solution in which all representations of the slots fall into one cluster, while still minimizing the objective in Equation \ref{eq:final_loss}. 

To avoid this degenerate case, we employ a similar set of ideas as DINO \cite{Caron2021EmergingPI} to have asymmetric teacher and student branches: 1) using data augmentations of teacher and student views; 2) centering and sharpening of teacher slot distributions; 3) updating teacher weights by taking an exponential moving average of student. Typically, the teacher model's weights are updated after every gradient update step for most datasets. However, for the CMNIST datasets, data augmentation is not used. To maintain the asymmetry between the teacher and student models in the absence of data augmentation, the updates for the CMNIST datasets are performed less frequently, specifically after every 20 steps.
\section{Conclusion}
\label{sec:conclusion}
Drawing inspiration from object-centric representation learning based on slot attention, we proposed a framework for decomposing images into concepts in an unsupervised way. We demonstrated the effectiveness of these concept clusters in discerning between minority and majority group samples within the dataset, all without relying on human group annotations. Leveraging these concepts, we devised an importance sampling technique that prioritizes rare concepts for each class, culminating in the training of a robust model exhibiting consistent performance with existing baselines in mitigating worst group errors. 

Our exploration in this paper has been confined to vision datasets; however, future investigations could extend to NLP or multi-modal datasets to further alleviate biased learning. Additionally, promising research avenues involve techniques targeting spurious concepts, such as concept-aware data augmentations, warranting further exploration.

\section*{Acknowledgements}
We acknowledge the support of the Canada CIFAR AI Chair Program and IVADO. We thank Mila and Compute Canada for providing computational resources.



\nocite{langley00}

\bibliography{main}
\bibliographystyle{icml2024}

\newpage
\appendix
\onecolumn
\section{Detailed architecture}
\label{app:detailed_arch}

\textbf{Encoders and Projectors.}
We employ a \emph{student encoder} $f_{s}$, \emph{projector} $g_{s}$ for a branch we call the \emph{student branch} which is updated using stochastic gradient descent (SGD). For simplicity, we represent the parameter weights of both the encoder and the projector of the student branch by $\theta_s$.
Similarly, the other branch named the \emph{teacher branch}, parameterised by $\theta_t$, has the same set of architectural components, respectively, \emph{teacher encoder} $f_{t}$, \emph{projector} $g_{t}$, but with different sets of weights, which are updated by the exponential moving average of the student parameters as:

For an input image, two randomly augmented views $x_s, x_t \in R^{c\times h\times w}$ are created, where $c$, $h$, and $w$ are input image channel, height and width respectively;  the goal is to extract slot representations from one view of the teacher branch and apply consistency to another view of the student branch.
Firstly, the augmented views are encoded and projected by the student and teacher encoder and projectors as 
$y_{s}=g_s(f_{s}(x_s)) \in R^{P \times d}$ and $y_{t}=g_{t}(f_{t}(x_t)) \in R^{P \times d}$, where are $P$ spatial patch representations of dimension $d$.  

\textbf{Slot Module.}
We can learn abstract concept representations by grouping semantically similar patch representation vectors. To do that, we employ randomly initialized vectors for the student branch, which we call \emph{student slots} $S_s \in R^{N\times d}$, where $N$ is the number of slots with dimension $d$. These slot vectors are then used to perform an attention-weighted pooling of patch vectors. To encourage competition among slots, where each slot attends to distinctive and semantically similar patch vectors, we utilize slot attention~\cite{locatello2020object} where attention is calculated over the `slot' axis as below:

\emph{We define any normalized vector as $\bar{z} = z / ||z||$ and any column normalized matrix as $\bar{Z} = [\bar{z}_1, ..., \bar{z}_N]$}

\begin{align}
    A_{s}&=\softmax_N(\bar{S_s}.\bar{y_{s}}^T/\tau_s) \in R^{N \times P}
\end{align}
where $\tau_s$ is the temperature for the student.

Then we can calculate \textbf{concept representations} using slots by pooling the patch representations based on the attention maps as follows:
\begin{equation}
    \begin{split}
    z_{s} = A_{s} . y_{s}, \quad z_{t} = A_{t} . y_{t} \in R^{N \times d}
    \end{split}
\end{equation}

\paragraph{Teacher}

We do the same for the teacher branch, \emph{teacher slots} by initializing them from the student slots weights $S_t \in R^{N\times d}$ as:
The weights of $S_s$ are updated by SGD, but the weights of $S_t$ are updated by the exponential moving average of $S_s$ as follows:
\begin{align}
    \theta_t &= \alpha_t \cdot \theta_s + (1 - \alpha_t) \cdot \theta_t\\
    S_t &= (1 - \alpha_t) \cdot S_s +  \alpha_t \cdot S_t\\
    A_{t} &= \softmax_N(\bar{S_t}.\bar{y_{t}}^T/\tau_t) \in R^{N \times P}
\end{align}
where $\tau_t$ is the temperature for the teacher.

\paragraph{Loss}
\label{app:loss_wen_el}
Since we utilize a two-branch teacher-student architecture for better inductive bias to facilitate higher-level abstraction, we focus on attention distillation from teacher to student. The student model learns to mimic the attention patterns of the teacher, effectively capturing the representation of the essential abstract concept from the data without explicitly reconstructing the input. 

We utilize the attention distillation loss introduced in~\citet{wen2022self}:
\begin{equation}
    \boldsymbol{\mathcal{L}_{dis}} = -\sum_N\sum_P M \circ A_{t} \log A_{s} 
\end{equation}
where $M$ is a mask that prevents distillation of non-overlapping patches from the teacher to the student views. Since the student and teacher branches observe two randomly cropped views, there may exist non-overlapping patches to which we do not want to apply distillation.

To avoid redundant slots and facilitate the learning of different information, we use a contrastive loss between the slot representations introduced in~\citet{wen2022self}. This ensures that similar concepts have closer representations in the embedding space, while dissimilar ones are further apart.

\begin{align}
    \boldsymbol{\mathcal{L}_{con}} &= \frac{1}{N}\sum_{i=1}^N -\log \frac{I(i) \exp \left(p(\bar{z}_{s}^i). \bar{z}_{t}^i/\tau_c \right)}{\sum_{\hat{i}^=1}^N I(\hat{i}) \exp \left(p(\bar{z}_{s}^{\hat{i}}), \bar{z}_{t}^{\hat{i}}/\tau_c \right )}
\end{align}
where $p$ is predictor network as in~\cite{Caron2021EmergingPI} and $I$ is an indicator function that finds common slots between the views of the teacher and the student after masking out the slots that fail to attend to any patch as below:

\[
\mathbf{I}(i) = 
\begin{cases} 
1 & \text{if } (\mathbf{m}_{s})_i = (\mathbf{m}_{t})_i \\
0 & \text{otherwise}
\end{cases}
\label{eq:indicator}
\]

\[
(\mathbf{m}_{s})_i =  \sum_{j=1}^{P} \mathbbm{1}\{ i = {\text{argmax}} \, A_s^{:j} \} \geq 1
\label{eq:act_slot}
\]

\[
(\mathbf{m}_{t})_i =  \sum_{j=1}^{P} \mathbbm{1}\{ i = {\text{argmax}} \, A_t^{:j} \} \geq 1 
\]

\section{Implementation Details}

\subsection*{Architecture}
For the training of the concept learning model, Imagenet pre-trained ResNet-50 has been used for the student encoder $f_s$ and the teacher encoder $f_t$. The student $p_s$ and teacher $p_t$ projector networks are similar to\cite{Caron2021EmergingPI} with a hidden dimension of $1024$ and an output dimension of $32$. We also use $32$ as the slot dimension and the concept vector dimension for all datasets except CMNIST. For CMNIST, we use slot and hidden dimensions of $16$ and $32$, respectively.

\subsection{Training Details} 
We train the concept discovery and classification model using ResNet50~\cite{he2016deep} pre-trained on ImageNet from the Pytorch library~\cite{paszke2019pytorch} as a backbone for all data sets except CMNIST where LeNet-5 \citep{lecun1998gradient} is used.  All experiments were performed with NVIDIA A100 and V100 GPUs.

\subsection*{Data Augmentation}
For the training of the concept learning model, we follow the data augmentation scheme proposed in~\citet{wen2022self}.

\subsection*{Hyperparameters} For both the student and teacher networks, the temperature values $\tau_s$ and $\tau_t$ are $0.1$ and $0.07$, similar to~\citet{Caron2021EmergingPI}. For the exponential moving average (EMA) update coefficients $\alpha_c$ and $\alpha_t$, we use $0.9$ and $0.99$ respectively. To obtain the sampling factor ($\lambda$) in \cref{alg:batch-sampling}, we use the average silhouette score as in~\citet{yang2024identifying}. For all datasets, when the average silhouette score $\le0.8$, we set $\lambda=2$, otherwise $1$. For Waterbirds and Urban Cars, it is $2$, and for other data sets, it is set to $1$.

For training the concept discovery model, we use Adam~\cite{kingma2017adam} as an optimizer with a learning rate of $2e^{-4}$ and a weight decay of $5e^{-4}$ for $50$ epochs with a batch size of $128$. The same configuration is used for all data sets, except CMNIST and CelebA, which are trained for $20$ epochs. For CMNIST, the batch size is $32$.

We train classification models using SGD with $0.9$ momentum for all datasets. The learning rate is $1e^{-4}$, except for CMNIST ($1e^{-3}$). A weight decay of $0.1$ is applied to Waterbirds, CelebA, and Urban Cars. Training epochs: Waterbirds ($300$), CelebA ($60$), IN-9L ($100$), 
Urban Cars ($300$), and CMNIST ($20$). The batch size is $128$ for all datasets, except CMNIST ($32$).

\section{Additional Results}
\label{app:addional_result}
\subsection*{Ablation Studies}
\begin{figure*}[htbp]
   \centering
   \includegraphics[width=.89\textwidth]{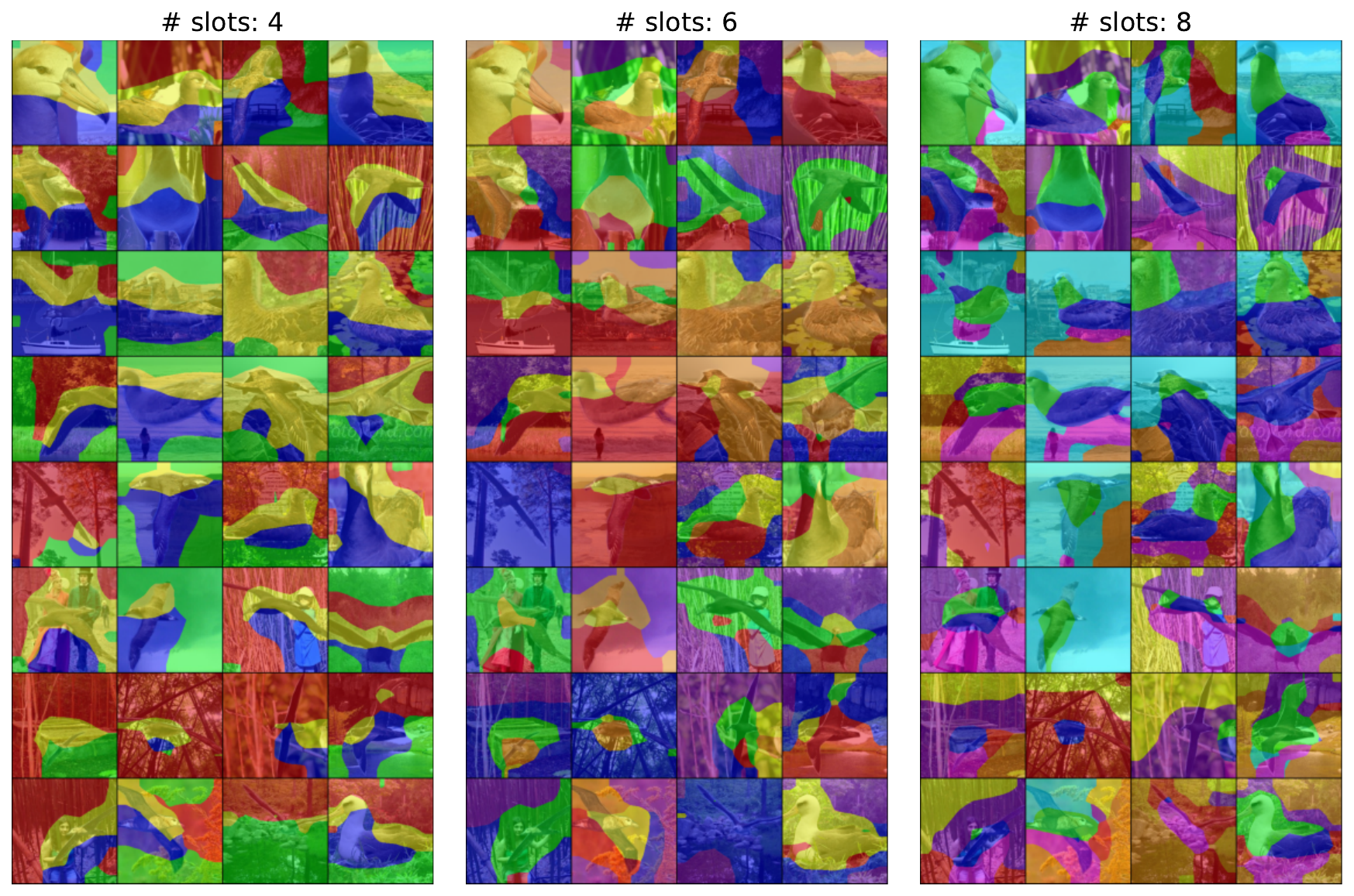}
   \caption{Ablation of segmentation mask with varying slots on Waterbirds}
   \label{fig:seg_mask_ablation}
\end{figure*}

\begin{figure*}[htbp]
   \centering
   \includegraphics[width=.89\textwidth]{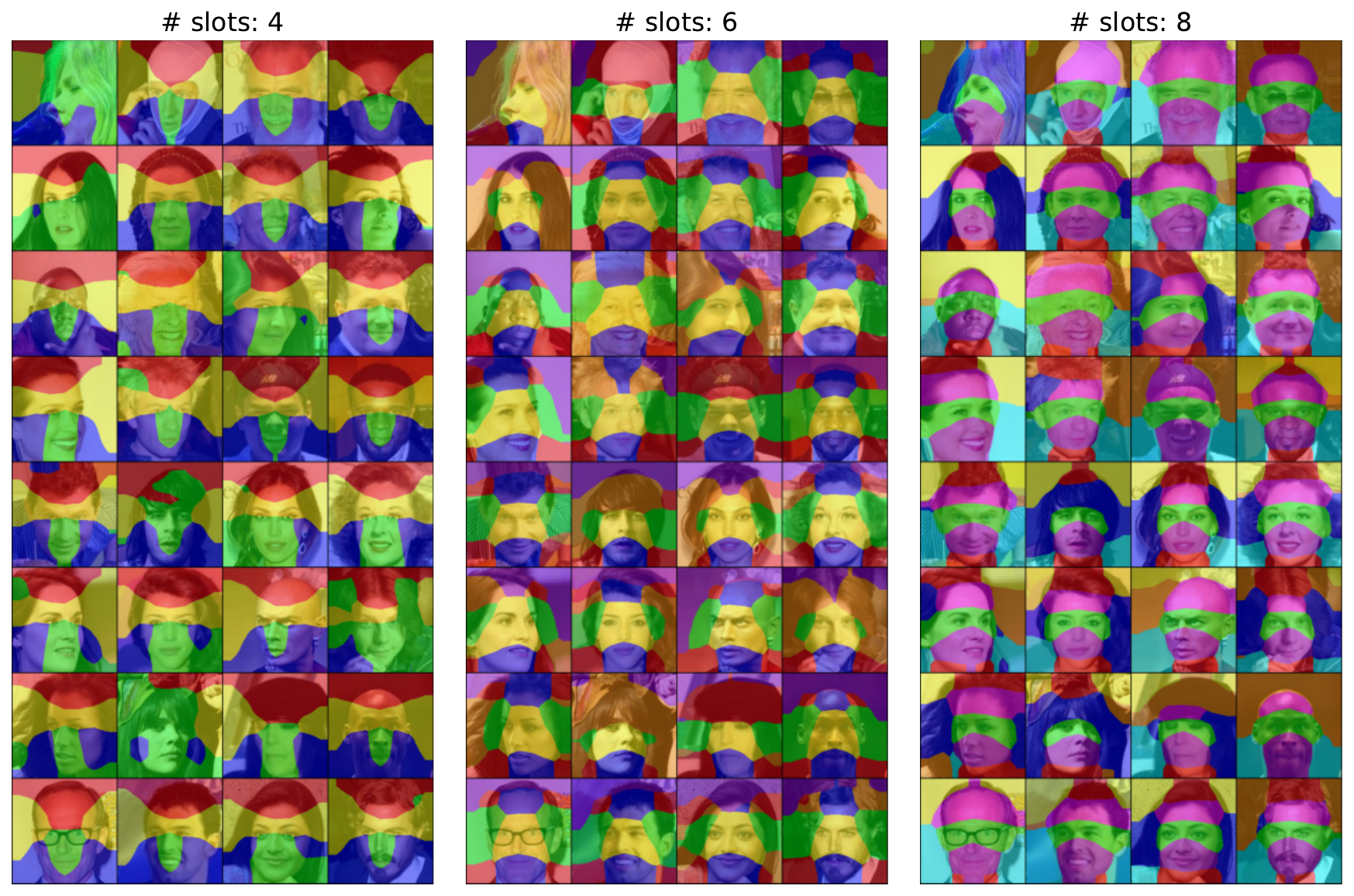}
   \caption{Ablation of segmentation mask with varying slots on CelebA}
   \label{fig:seg_mask_ablation_cb}
\end{figure*}

\begin{figure*}[htbp]
   \centering
   \includegraphics[width=.89\textwidth]{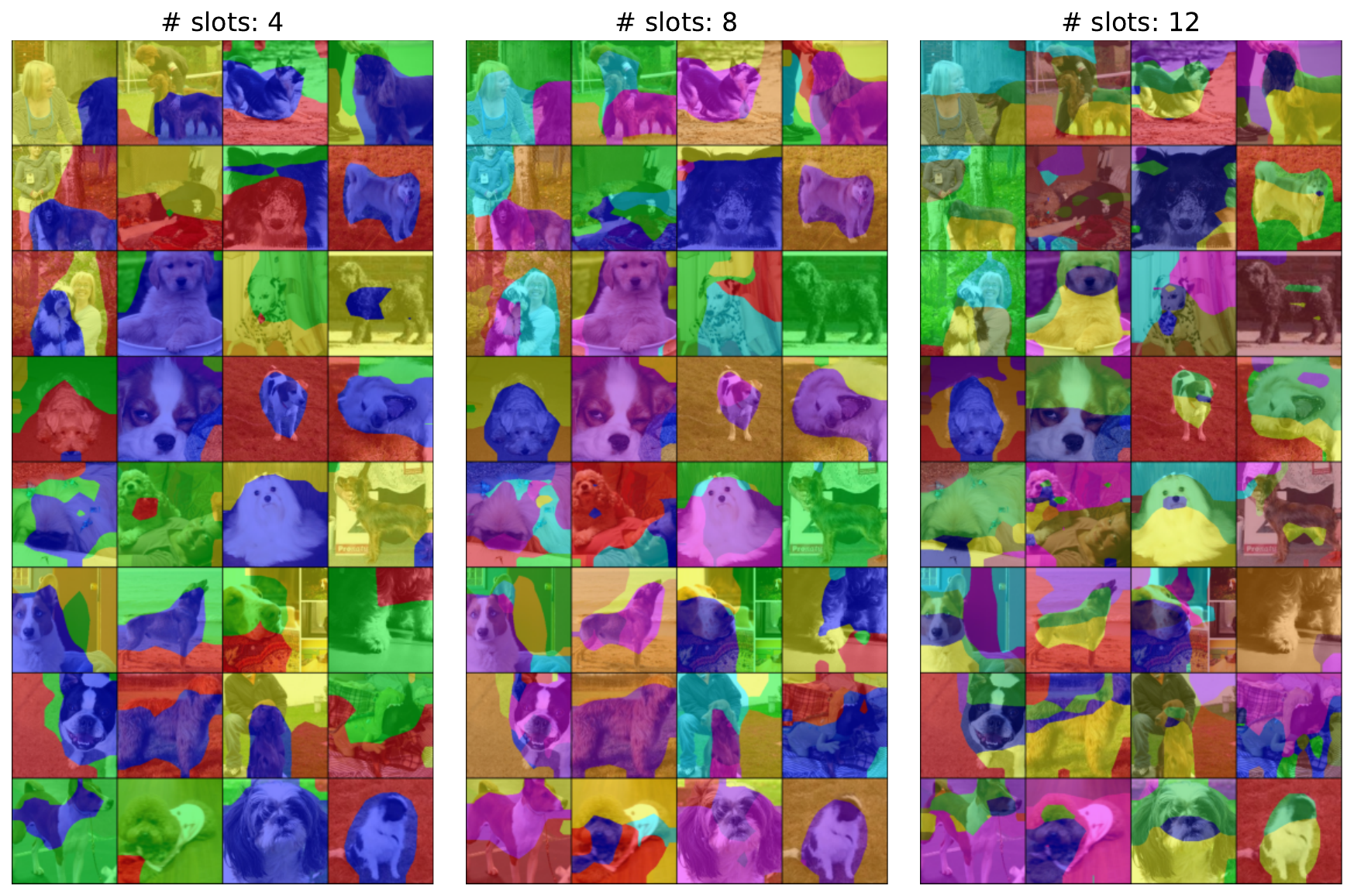}
   \caption{Ablation of segmentation mask with varying slots on IN-9L}
   \label{fig:seg_mask_ablation_in9l}
\end{figure*}

To perform an ablation study on the \textbf{Waterbirds} dataset, we used different numbers of slots $\{2, 4, 6, 8\}$ and the size of the codebook $\{2, 4, 8, 12, 16\}$. From~\cref{fig:seg_mask_ablation}, we can see the segmentation mask with a varying number of slots. From the figure, it is clear that on average $3-4$ and $2-6$ slots are activated per image when we initialize with $4$ and $6$ slots, respectively. We can identify the activated slots for each image based on~\autoref{eq:act_slot}. If we increase the number of slots, we can see fine-grained segmentation. ~\cref{fig:seg_mask_ablation_cb} and ~\cref{fig:seg_mask_ablation_in9l} shows similar ablation studies of segmented images while varying the $\#$ slots in the \textbf{CelebA} and \textbf{IN-9L} data sets, respectively.

\begin{figure}[h]
\centering
  \centering
  \includegraphics[scale=0.27]{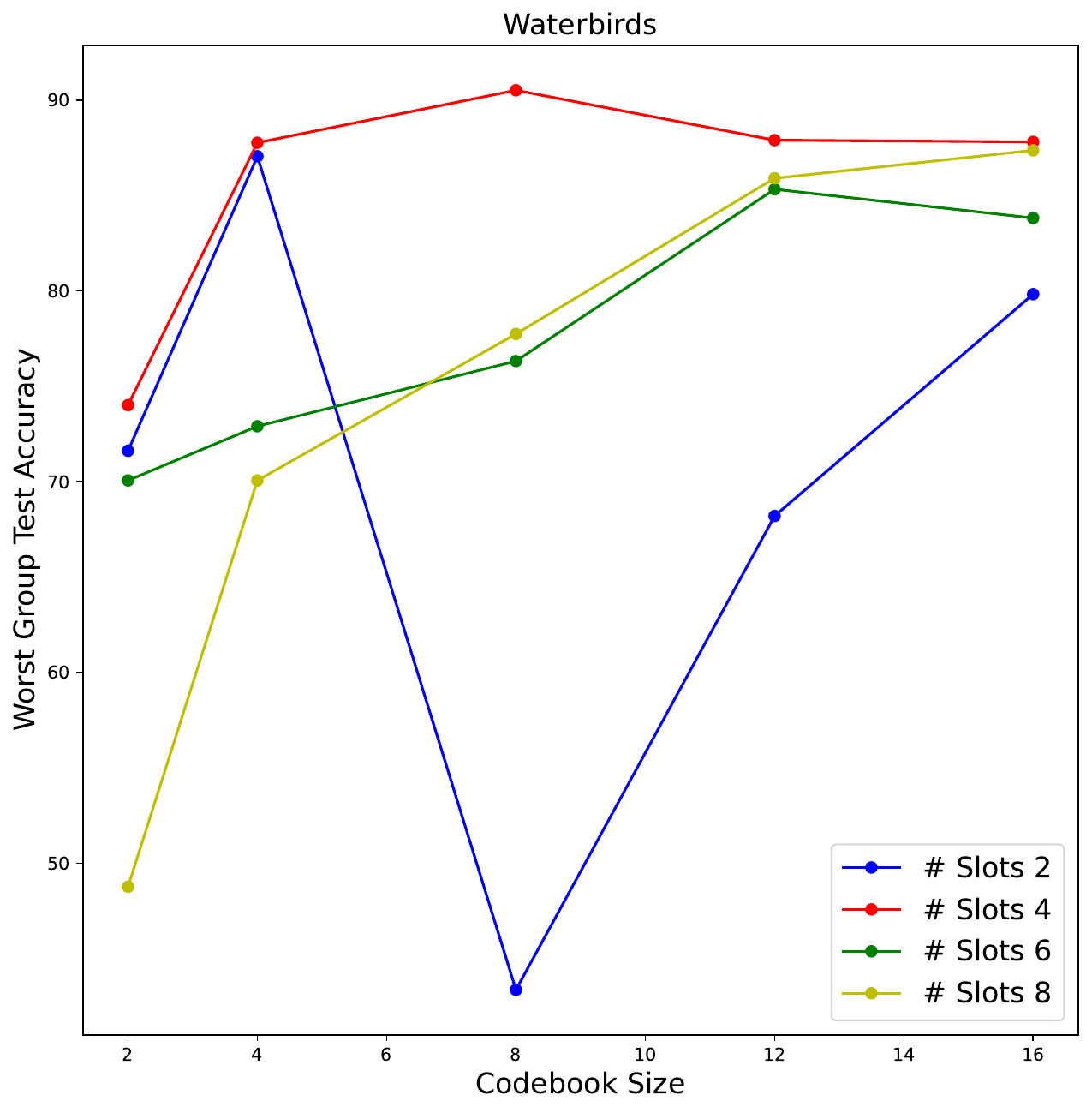}
\caption{Ablation on Worst group  Test Accuracy on Waterbirds dataset with varying slots and codebook size}
\label{fig:wb_acc_ablation}
\end{figure}

From~\cref{fig:wb_acc_ablation}, we can see the impact on the performance of the worst group accuracy when varying the number of slots and the size of the codebook. It is evident that over-segmentation (with more slots) degrades the worst group accuracy. We hypothesize that we can have the right balance between the worst group and the average accuracy with a suitable number of slots, which encourages us to learn abstract semantic concepts. In all of our experiments, we use $4$ as the $\#$ slot for all data sets. 

 \cref{table:ablation_celeba_acc} shows the worst group and average accuracy on \textbf{CelebA} dataset while varying the size of the codebook and fixing the $\#$ slots to $4$. 

\begin{table}[h]
\centering
\begin{tabular}{c|c|c}
\hline
Codebook Size & Worst group Acc& Avg Acc \\ \hline
4 & 74.4 & 93.5  \\ 
8 & 88.3 & 92.6  \\ 
12 & 86.7 & 89.7  \\ 
16 & 88.3 & 90.4 \\ 
\end{tabular}
\caption{Varying Codebook size on CelebA ($\#$ slots fixed to $4$)}
\label{table:ablation_celeba_acc}
\end{table}

\subsection*{Results on Bar Dataset with Spatially Separable Concepts}
We further evaluated our method on the Biased Action Recognition (BAR) dataset \citep{nam2020learning}, which includes six action classes biased to distinct places. In this dataset, the concepts have clear spatial regions and we see (Table \ref{tab:bar}) the efficacy of our method, where it outperforms other baselines.
\begin{table}[h]
  \centering
  \caption{Result on BAR \citep{nam2020learning} dataset}
  \begin{tabular}{ccccc}
    \toprule
    ERM & ReBias \citep{bahng2020learning} & LfF & BiaSwap \citep{kim2021biaswap} & \SLOT{}$_{ig}$ \\ \hline
    51.9±5.9	& 59.7±1.5 &	63.0±2.8 & 52.4	& \textbf{67.0±0.9}
   \\ 
   \bottomrule
  \end{tabular}
  \label{tab:bar}
\end{table}

\end{document}